\documentclass[review]{elsarticle}

\usepackage{hyperref}
\usepackage[utf8]{inputenc}
\usepackage{bm}
\usepackage{amsmath}
\newcommand\norm[1]{\left\lVert#1\right\rVert}
\usepackage{algpseudocode}
\usepackage{algorithm}
\usepackage{mathtools}
\usepackage{amsmath}
\usepackage{amssymb}
\usepackage{xcolor}
\usepackage{rotating}

\usepackage{amsthm}
\usepackage{enumitem}
  \newtheoremstyle{dotless}{}{}{\itshape}{}{\bfseries}{}{ }{}
  \theoremstyle{dotless}
  \newtheorem{thm}{Theorem}

\journal{}

\begin{document}

\begin{frontmatter}

\title{$\ell_p$ Slack Norm Support Vector Data Description}

\author{Shervin Rahimzadeh Arashloo\fnref{myfootnote}}
\address{}
\fntext[myfootnote]{Corresponding author: Shervin Rahimzadeh Arashloo.\\ E-mail: s.rahimzadeh@surrey.ac.uk}


\begin{abstract}
The support vector data description (SVDD) approach serves as a de facto standard for one-class classification where the learning task entails inferring the smallest hyper-sphere to enclose target objects while linearly penalising any errors/slacks via an $\ell_1$-norm penalty term. In this study, we generalise this modelling formalism to a general $\ell_p$-norm ($p\geq1$) slack penalty function. By virtue of an $\ell_p$ slack norm, the proposed approach enables formulating a non-linear cost function with respect to slacks. From a dual problem perspective, the proposed method introduces a sparsity-inducing dual norm into the objective function, and thus, possesses a higher capacity to tune into the inherent sparsity of the problem for enhanced descriptive capability.

A theoretical analysis based on Rademacher complexities characterises the generalisation performance of the proposed approach in terms of parameter $p$ while the experimental results on several datasets confirm the merits of the proposed method compared to other alternatives.
\end{abstract}
\end{frontmatter}


\section{Introduction}
One-class classification (OCC) addresses the problem of recognising patterns that adhere to a specific condition presumed as normal, and identifying them from any other object violating the normality criterion. OCC stands apart from the conventional two-/multi-class classification paradigm in that it primarily uses observations from a single, very often the target class for training. One-class classification acts as an essential building block in a diverse range of practical systems including presentation attack detection in biometrics \cite{FATEMIFAR2021107696}, health care \cite{CARRERA2019482}, audio or video surveillance \cite{4668357,ZHANG2020107394}, intrusion detection \cite{6846360}, social network \cite{CHAKER2017266}, safety-critical systems \cite{4694106}, fraud detection \cite{Kamaruddin:2016:CCF:2980258.2980319}, insurance \cite{7435726}, etc.

As with many other machine learning problems, state-of-the-art OCC algorithms are built on the premise of deep learning methodology \cite{Goodfellow-et-al-2016,pmlr-v80-ruff18a,ERFANI2016121} using massive labelled datasets, typically containing millions of samples. Although deep structures have led to breakthroughs in one-class learning and classification, their reliance on huge sets of data may pose certain limitations in practice. In this context, collecting sufficiently large sets of training observations for certain applications can be a challenge, hindering a full exploitation of the expressive capacity of deep networks. Even if sufficient data is gathered, labelling such huge amounts of data may be a bigger challenge. Whilst crowd-sourcing may be considered as an applicable strategy to label huge sets of data in some fields, for a variety of different reasons including level of knowledge, data privacy, time required to produce accurate labels, etc. it may not serve as a viable option in the domains such as defence, security, healthcare. Although certain techniques such as active learning \cite{series/synthesis/2012Settles} or learning with privileged information \cite{JMLR:v16:vapnik15b} may be instrumental in reducing the quantity of necessary labelling/labelled data, they still demand the time and domain expertise of a human operator. In the absence of large training sets required by deep nets, and specifically for small to moderate-sized datasets containing hundreds or thousands of training samples, kernel-based methods \cite{Cortes1995} offer a very promising methodology of classification. Moreover, unlike deep networks that incorporate many heuristics with regards to their structure and the corresponding (hyper)parameters, kernel methods are based on solid foundations and are characterised by strong bases in optimisation and statistical learning theory \cite{Vapnik1998}.

The support vector data description (SVDD) approach \cite{Tax2004} which is proposed as an adaptation of support vector machines to the one-class setting, presents a very popular kernel-based method for one-class classification. Although designed for one-class setting, the SVDD approach does not require the training data to be exclusively and purely normal/positive which can be regarded as a quite appealing property in practical applications where the data is very often contaminated with noise and outliers. Furthermore, it provides an intuitive geometric characterisation of a predominantly positive dataset without making any specific assumption regarding the underlying distribution. Moreover, the SVDD decision making process entails computing a simple distance between the centre of the target class and a test observation to label it as either normal (i.e. positive/target) or as anomaly (i.e. negative/outlier). And finally, when large sets of training data are available, the SVDD method may be extended to a deep structure to directly learn features from the data for improved performance \cite{pmlr-v80-ruff18a}. These properties make SVDD a highly favoured method of practice in a variety of one-class classification applications where it serves as one of the most widely used techniques, if not the most.

The underlying idea in the SVDD approach is to determine the smallest hyper-sphere enclosing the data. While its hard-margin formulation requires all positive target data to be strictly encapsulated within the inferred hyper-spherical boundary, in practical situations, a dataset may incorporate noisy/outlier samples. In the soft margin SVDD approach, in order to take into account the possibility of a contaminated dataset and improve the generalisation capability of the model, the distance from each training object to the centre of the hyper-sphere need not be strictly smaller than the radius but larger distances are penalised. In order to encode and penalise violations from the hyper-spherical decision boundary, in the soft margin variant, non-negative slack variables measuring the extent of violation of each object from the decision boundary are introduced. The optimisation problem is then modified to reflect such violations and penalise an $\ell_1$-norm term on the slacks. In other words, the conventional SVDD method, and also the standard two-class SVM classifier \cite{Cortes1995} are founded on the idea of minimising an $\ell_1$-norm risk over the set of non-negative slack variables. In the context of two-class classification, very recently \cite{VAPNIK2021108018}, the classical $\ell_1$-norm penalty term in the SVM formulation has been revisited to consider two alternative slack penalties defined by the $\ell_2$- and the $\ell_\infty$-norms to formulate new SVM algorithms. A reformulation of the standard two-class SVM to the $\ell_2$ and $\ell_\infty$ penalty terms has been verified to improve the classification performance, sometimes significantly \cite{VAPNIK2021108018}.

In this work, we study the merits of different norm risks for "one-class" classification in the context of the SVDD approach. For this purpose, we consider a general $\ell_p$-norm ($p\geq1$) slack penalty term where $p$ serves as a free parameter of the algorithm. As such, while in the standard SVDD method the slacks are penalised linearly, by introducing a $p$-norm function, non-linear cost functions of the slacks may be optimised where the degree of the non-linearity (i.e. $p$) may be tuned on the data. The reflection of the $\ell_p$ slack penalty term onto the dual space formulation of the corresponding optimisation problem turns out to be a dual $\ell_q$-norm ($q=\frac{p}{p-1}$) cost on the dual variables, thus, providing the method the capability to tune into the inherent sparsity of the problem.

\subsection{Contributions}
The major contributions of the current study may be summarised as listed below.
\begin{itemize}
    \item We generalise the SVDD formulation from an $\ell_1$ to an $\ell_p$ slack norm penalty function and illustrate that the proposed generalisation may lead to significant improvements in the performance of the algorithm;
    \item We extend the proposed $\ell_p$-norm formulation from a pure one-class setting to the training scenario where labelled negative objects are also available and illustrate the merits offered by the proposed extension;
    \item Based on Rademacher complexities, we theoretically study the generalisation performance of the proposed $\ell_p$ slack norm approach and derive bounds on its error;
    \item And we carry out an experimental evaluation of the proposed method on multiple OCC datasets and provide a comparison to the original SVDD method and its different variants, as well as other OCC techniques from the literature.
\end{itemize}

\subsection{Organisation}
The remainder of the paper is structured as follows. In Section \ref{lit}, the relevant literature with a particular emphasis on different variants of the SVDD formalism is reviewed. In Section \ref{propp}, once a short overview of the support vector data description (SVDD) approach \cite{Tax2004} is given, we present our proposed $\ell_p$ slack norm SVDD approach for the pure one-class setting and then derive its extension for labelled negative training observations. Section \ref{geb} studies the generalisation error bound of the proposed approach based on Rademacher complexities. We present and analyse the results of an experimental evaluation of the proposed method in Section \ref{exps} where possible extensions of the proposed approach are also discussed. Finally, Section \ref{conc} concludes the paper.

\section{Prior work}
\label{lit}
Although different categorisations of OCC methods exist in different studies \cite{10.1145/1541880.1541882,khan_madden_2014,PIMENTEL2014215,DBLP:journals/corr/abs-2101-03064,Tax2004}, the one-class classification techniques may be roughly identified as either generative or non-generative \cite{6636290}, the latter best represented by discriminative approaches. While in the generative techniques, the objective is to model the underlying generative process of the data, the discriminative methods try to directly partition the observation space into different regions for classification. Discriminative approaches tend to yield better performance in practice since they try to explicitly solve the OCC problem without attempting to solve an intermediate and more general task of inferring the underlying distribution or generative process \cite{Vapnik1998}.

Generative OCC approaches encompass the methods that try to estimate the underlying distribution using, for example, Parzen windowing \cite{bish}, Gaussian distribution modelling \cite{6796568}, or those which use a mixture of distributions \cite{PlatzerDSNW08,FATEMIFAR2022108500}. A different sub-category of generative approaches include methods that for decision making use the residual of reconstructing a test sample with respect to a hypothesised model, some instances of which are the kernel principal component analysis (KPCA) and its variants \cite{HOFFMANN2007863,XIAO2013389}, or the autoencoder-based techniques \cite{8953541}.

Discriminative methods constitute a strong alternative to the generative one-class learners. As an instance, based on a variant of the Fisher classification principle, the kernel null space method \cite{6619277} tries to map positive objects onto a single point in a discriminative subspace, obtaining very competitive results compared to some other alternatives \cite{9055448}. Another successful discriminative one-class method focuses on the use of Gaussian Process (GP) priors \cite{KEMMLER20133507} trying to directly infer the a posteriori class probability of the target class. A further example of discriminative one-class learners is that of the nearest neighbour-based approaches \cite{906164} where the normality of an object is decided based on its immediate neighbours. Among others, a widely applied discriminative one-class classification method is that of support vector data description (SVDD) approach \cite{Tax2004} that tries to estimate the smallest volume surrounding the positive objects. In the case of the existence of labelled negative training objects, the decision boundary is refined by requiring the negative objects to lie outside the hyper-spherical boundary. The soft version of this approach allows the positive and negative (if any) training objects to violate the boundary criterion but subject to a linear penalty on the extent of the violation (called \textit{slack}) where a parameter controls the trade-off between the volume and such errors in the objective function. Due to its success in data description and its intuitive geometrical interpretation and the ability to benefit from a kernel-based representation, the SVDD approach serves as a widely used technique in the OCC literature, motivating many subsequent research. As an instance, in \cite{9632278}, based on the observation that the SVDD centre and the volume are sensitive to the parameter controlling the trade-off between the errors (slacks) and the volume, a method called GL-SVDD is proposed where local and global probability densities are used to derive sample-adaptive errors via associating weights to the slacks corresponding to different objects. In \cite{WANG2013875}, a different sample-specific weighting approach (P-SVDD) based on the position of the feature space image is proposed to adaptively regularise the complexity of the SVDD sphere. Other work \cite{CHA20143343} (DW-SVDD) considers re-weighting sample errors in the objective function by utilising the relative density of each object to the density distribution of normal samples. The authors in \cite{LEE20051768} define a density-based distance between a sample point and the centre of the hyper-sphere to adjust the constraint set of the SVDD optimisation problem by re-weighting training objects. The work in \cite{CHEN2015129}, considers a different linear sample weighting scheme in the SVDD objective function by introducing the cut-off distance-based local density of objects. The work in \cite{4760149} introduces a margin parameter to maximise the margin between the hyper-sphere and the non-target objects in an SVDD formulation and directly optimises the margin. The Euclidean distance ($\ell_2$ distance) employed in the widely used Gaussian kernel function is reassessed in \cite{6846360} to see if other distances in the Gaussian kernel function may provide performance advantages. Apart from the research focused on improving the performance of the SVDD method in a one-class setting, there also exist other studies where the SVDD approach is generalised to two \cite{HUANG2011320}, or to multiple classes \cite{TURKOZ2020107119}.

Considering the body of work discussed above, one observes that the majority of the existing studies tries to modify the slack error term by introducing an adaptive weighting for each data sample based on different cues. Clearly, a simple linear weighting scheme does change the linearity of the objective function with respect to the slacks. The exceptions to the studies above are the work in \cite{JMLR:v6:tsang05a, Chang_arevisit} where instead of an $\ell_1$-norm penalty, an $\ell_2$-norm penalty is considered over the slacks. As will be demonstrated in the subsequent sections, an $\ell_2$ slack penalty may not always yield an optimal performance for data description. The current study is a generalisation of the existing SVDD formulations as it considers an $\ell_p$ ($p\geq1$) slack norm penalty where $p$ serves as a free parameter of the algorithm allowing for different non-linear penalties to be optimised w.r.t. slacks while at the same time providing the opportunity to tune into the inherent sparsity characteristics of the data.  

\section{Methodology}
\label{propp}
In this section, first, we briefly review the SVDD method \cite{Tax2004} and then present the proposed approach.
\subsection{Preliminaries}
The Support Vector Data Description (SVDD) approach \cite{Tax2004} tries to estimate the smallest hyperspherical volume that encloses normal/target data in some pre-determined feature space. As a hypersphere is characterised by its centre $\mathcal{O}$ and its radius $R>0$, the learning problem in the SVDD method is defined as minimising the radius while requiring the hypersphere to encapsulate all normal objects $\mathbf{x}_i$'s, that is
\begin{align}
\nonumber &\min_{R,\mathcal{O}}E(R,\mathcal{O})=R^2\\
& \text{s.t. } \norm{\mathbf{x}_i-\mathcal{O}}_2^2\leq R^2, \text{ }\forall i
\end{align}
In practice, however, the training data might be contaminated with noise and outliers. In order to handle possible outliers in the training set and derive a solution with a better generalisation capability, the objective function in the SVDD method is modified so that the distance from the centre $\mathcal{O}$ to each training observation $\mathbf{x}_i$ need not be strictly smaller than $R$, rather, larger distances are penalised. For this purpose, using non-negative slack variables $\zeta_i$'s, the SVDD optimisation task is modified as
\begin{align}
\nonumber & \min_{R,\mathcal{O},\boldsymbol\zeta}E(R,\mathcal{O},\boldsymbol\zeta)=R^2+c\sum_i\zeta_i\\
& \text{s.t. } \norm{\mathbf{x}_i-\mathcal{O}}_2^2\leq R^2+\zeta_i, \text{  } \zeta_i\geq0, \text{  } \forall i
\label{sl1}
\end{align}
where $\boldsymbol\zeta$ denotes a vector collection of $\zeta_i$'s and the trade-off between the sum of errors (i.e. $\zeta_i$'s) and the squared radius is controlled using parameter $c$. The optimisation problem above corresponds to the case where only normal samples (and possibly a minority of noisy objects) are presumed to exist in the training set. When labelled negative training objects are also available, the learning problem in the SVDD method is modified to enforce positive samples to lie within the hyper-sphere while negative samples are encouraged to fall outside its boundary.

The SVDD objective function in Eq. \ref{sl1} depends on an $\ell_1$-norm of the slack variables as $\sum_i\zeta_i=\norm{\boldsymbol\zeta}_1$, and consequently, all errors/slacks are penalised linearly. Although penalising all errors linearly in their magnitudes is a
plausible option, it is by no means the only possibly. An an instance, a different alternative may be to penalise only the maximum error/slack which can be achieved by incorporating an $\ell_{\infty}$-norm on the slacks as $\max_i\zeta_i=\norm{\boldsymbol\zeta}_{\infty}$. Any other penalty which would lie between penalising all the slacks linearly and penalising only the maximum error may then be characterised using a general $\ell_p$-norm on the errors, i.e. via $\sum_i\zeta_i^p=\norm{\boldsymbol\zeta}_p^p$. In particular, introducing a variable norm parameter $p$ opens the door to consider non-linear penalties on the errors compared with the original SVDD method which is limited to a linear penalty on the slacks. From a dual problem viewpoint, introducing an $\ell_p$ norm penalty on the slacks translates into sparsity inducing dual norms on the dual variables which provides the opportunity to better consider the intrinsic sparsity of the problem. As such, in the proposed approach, we generalise the SVDD error function using an $\ell_p$-norm function of slacks, discussed next.
\subsection{$\ell_p$ slack norm SVDD}
By replacing the $\ell_1$-norm term on the slack variables in Eq. \ref{sl1} with an $\ell_p$-norm, the optimisation problem in the proposed approach is defined as
\begin{align}
\nonumber & \min_{R,\mathcal{O},\boldsymbol\zeta}E(R,\mathcal{O})=R^2+c\sum_i\zeta_i^p\\
& \text{s.t. } \norm{\mathbf{x}_i-\mathcal{O}}_2^2\leq R^2+\zeta_i, \text{  } \zeta_i\geq0, \text{  } \forall i
\label{sl2}
\end{align}
In order to solve the optimisation problem above, the Lagrangian is formed as
\begin{align}
    \mathcal{L}=R^2+c\sum_i\zeta_i^p-\sum_i\alpha_i[R^2+\zeta_i-(\norm{\mathbf{x}_i}_2^2-2\mathcal{O}^\top\mathbf{x}_i+\norm{\mathcal{O}}_2^2)]-\sum_i\gamma_i\zeta_i
\end{align}
where $\alpha_i$'s and $\gamma_i$'s are non-negative Lagrange multipliers. In order to derive the dual function, the Lagrangian should be minimised with respect to the primal variables $R$, $\mathcal{O}$, $\zeta_i$. Setting the partial derivatives of $\mathcal{L}$ w.r.t. $R$, $\mathcal{O}$, and $\zeta_i$ to zero yields
\begin{subequations}
\begin{align}
     &\frac{\partial \mathcal{L}}{\partial R}=0 \hspace{10pt} \Rightarrow \hspace{10pt} \sum_i\alpha_i=1\\
     &\frac{\partial \mathcal{L}}{\partial \mathcal{O}}=0 \hspace{10pt} \Rightarrow \hspace{10pt} \mathcal{O}=\sum_i\alpha_i\mathbf{x}_i\\
    &\frac{\partial \mathcal{L}}{\partial\zeta_i}=0 \hspace{10pt} \Rightarrow \zeta_i=(\frac{\alpha_i+\gamma_i}{cp})^{\frac{1}{p-1}}
    \label{eq:subeq1}
\end{align}
\end{subequations}
Substituting the relations above into $\mathcal{L}$, the Lagrangian is obtained as
\begin{align}
    \mathcal{L}=(cp)^{\frac{-1}{p-1}}(1/p-1)\norm{\boldsymbol\alpha+\boldsymbol\gamma}_{p/(p-1)}^{p/(p-1)}+\sum_i\alpha_i\mathbf{x}_i^\top\mathbf{x}_i-\sum_i\sum_j\alpha_i\alpha_j\mathbf{x}_i^\top\mathbf{x}_j
    \label{L11}
\end{align}
where $\boldsymbol\alpha$ and $\boldsymbol\gamma$ denote vector collections of $\alpha_i$'s and $\gamma_i$'s. Furthermore, one can easily check that the Slater's condition is satisfied, and thus, the following complementary conditions also hold at the optimum:
\begin{subequations}
\begin{align}
    &\gamma_i\zeta_i=0,\forall i\label{eq:subeq2}\\
    &\alpha_i(R^2+\zeta_i-\norm{\mathbf{x}_i-\mathcal{O}}_2^2)=0, \forall i \label{null}
\end{align}
\end{subequations}
Using Eq. \ref{eq:subeq1} and Eq. \ref{eq:subeq2}, it must hold that $\gamma_i(\frac{\alpha_i+\gamma_i}{cp})^{\frac{1}{p-1}}=0$. Since $\alpha_i\geq0$ and $\gamma_i\geq0$, one concludes that $\gamma_i=0$, $\forall i$. As a result, the Lagrangian in Eq. \ref{L11} would be simplified as
\begin{align}
    \mathcal{L}=(cp)^{\frac{-1}{p-1}}(1/p-1)\norm{\boldsymbol\alpha}_{p/(p-1)}^{p/(p-1)}+\sum_i\alpha_i\mathbf{x}_i^\top\mathbf{x}_i-\sum_i\sum_j\alpha_i\alpha_j\mathbf{x}_i^\top\mathbf{x}_j
    \label{L1}
\end{align}
The dual problem entails maximising $\mathcal{L}$ in $\boldsymbol\alpha$:
\begin{align}
    \nonumber &\max_{\boldsymbol\alpha} \hspace{10pt}\mathcal{L}\\
    &\text{s.t. } \boldsymbol\alpha\geq0, \norm{\boldsymbol\alpha}_1=1
    \label{optL1}
\end{align}
Note that, for $p\geq1$, we have $p/(p-1)\geq1$, and consequently, the term $\norm{\boldsymbol\alpha}_{p/(p-1)}^{p/(p-1)}$ in the Lagrangian is convex w.r.t. $\boldsymbol\alpha$. Note also that the other terms in $\mathcal{L}$ are either linear or quadratic functions of $\boldsymbol\alpha$, and hence, are convex while the constraints are affine. As a result, the optimisation problem above is a convex optimisation task.

\subsection{$\ell_p$ slack norm SVDD with negative samples}
In the proposed $\ell_p$ slack norm approach, similar to the original SVDD method \cite{Tax2004}, when labelled non-target/negative training observations are available, they may be utilised to refine the description. In this case, as opposed to the positive samples that should be enclosed within the hypersphere, the non-target objects should lie outside its boundary. In what follows, the normal/positive samples are indexed by $i$, $j$ and the negative objects by $l$, $m$. In order to allow for possible errors in both the positive and the negative training samples, slack variables $\zeta_i$'s and $\zeta_l$'s are introduced. The optimisation problem when labelled negative samples are available is then defined as
\begin{align}
\nonumber & \min_{R,\mathcal{O},\boldsymbol\zeta}E(R,\mathcal{O},\boldsymbol\zeta)=R^2+c_1\sum_i\zeta_i^p+c_2\sum_l\zeta_l^p\\
& \text{s.t. } \norm{\mathbf{x}_i-\mathcal{O}}_2^2\leq R^2+\zeta_i, \text{     }\norm{\mathbf{x}_l-\mathcal{O}}_2^2\geq R^2-\zeta_l, \text{    }\zeta_i\geq0,\text{  }\zeta_l\geq0, \text{    } \forall i,l
\label{NTO}
\end{align}
In the objective function above, while $c_1$ may be used to control the fraction of positive training objects that fall outside the hypersphere boundary, $c_2$ may be adjusted to regulate the fraction of negative training samples that will lie within the hypersphere. By introducing Lagrange multipliers $\alpha_i\geq0$, $\alpha_l\geq0$, $\gamma_i\geq0$, $\gamma_l\geq0$, the Lagrangian of Eq. \ref{NTO} is formed as
\begin{align}
    \nonumber&\mathcal{L}=R^2+c_1\sum_i\zeta_i^p+c_2\sum_l\zeta_l^p-\sum_i\gamma_i\zeta_i-\sum_l\gamma_l\zeta_l\\
    &\nonumber -\sum_i\alpha_i[R^2+\zeta_i-(\norm{\mathbf{x}_i}_2^2-2\mathcal{O}^\top\mathbf{x}_i+\norm{\mathcal{O}}_2^2)]\\
    &-\sum_l\alpha_l[(\norm{\mathbf{x}_l}_2^2-2\mathcal{O}^\top\mathbf{x}_l+\norm{\mathcal{O}}_2^2)-R^2+\zeta_l]
    \label{L21}
\end{align}
In order to form the dual function, the Lagrangian should be minimised w.r.t. $R$, $\mathcal{O}$, $\zeta_i$'s, and $\zeta_l$'s. Setting the partial derivatives of $\mathcal{L}$ w.r.t. to $R$, $\mathcal{O}$, $\zeta_i$, and $\zeta_l$ to zero yields
\begin{subequations}
\begin{align}
     &\frac{\partial \mathcal{L}}{\partial R}=0 \hspace{10pt} \Rightarrow \hspace{10pt} \sum_i\alpha_i-\sum_l\alpha_l=1\\
     &\frac{\partial \mathcal{L}}{\partial \mathcal{O}}=0 \hspace{10pt} \Rightarrow \hspace{10pt} \mathcal{O}=\sum_i\alpha_i\mathbf{x}_i-\sum_l\alpha_l\mathbf{x}_l\\
     &\frac{\partial \mathcal{L}}{\partial\zeta_i}=0 \hspace{10pt} \Rightarrow \zeta_i=(\frac{\alpha_i+\gamma_i}{c_1p})^{\frac{1}{p-1}}\label{subeq3}\\
    &\frac{\partial \mathcal{L}}{\partial\zeta_l}=0 \hspace{10pt} \Rightarrow \zeta_l=(\frac{\alpha_l+\gamma_l}{c_2p})^{\frac{1}{p-1}}\label{subeq5}
\end{align}
\end{subequations}
Resubstituting the relations above into Eq. \ref{L21} gives
\begin{align}
    \nonumber \mathcal{L}&=(c_1p)^{\frac{-1}{p-1}}(1/p-1)\norm{\boldsymbol\alpha_T+\boldsymbol\gamma_T}_{p/(p-1)}^{p/(p-1)}+(c_2p)^{\frac{-1}{p-1}}(1/p-1)\norm{\boldsymbol\alpha_N+\boldsymbol\gamma_N}_{p/(p-1)}^{p/(p-1)}\\
    \nonumber&+\sum_i\alpha_i\mathbf{x}_i^\top\mathbf{x}_i-\sum_l\alpha_l\mathbf{x}_l^\top\mathbf{x}_l-\sum_i\sum_j\alpha_i\alpha_j\mathbf{x}_i^\top\mathbf{x}_j-\sum_l\sum_m\alpha_l\alpha_m\mathbf{x}_l^\top\mathbf{x}_m\\
    &+2\sum_i\sum_l\alpha_i\alpha_l\mathbf{x}_i^\top\mathbf{x}_l
    \label{L2}
\end{align}
where $\boldsymbol\alpha_T$ and $\boldsymbol\alpha_N$ respectively stand for vector collections of $\alpha_i$'s and $\alpha_l$'s. Similarly, $\boldsymbol\gamma_T$ and $\boldsymbol\gamma_N$ denote vector collections of $\gamma_i$'s and $\gamma_l$'s, respectively. Since the Slater's condition holds, the following complementary conditions are also satisfied at the optimum:
\begin{subequations}
\begin{align}
    &\gamma_i\zeta_i=0,\forall i\label{subeq4}\\
    &\gamma_l\zeta_l=0,\forall l\label{subeq6}\\
    &\alpha_i(R^2+\zeta_i-\norm{\mathbf{x}_i-\mathcal{O}}_2^2)=0, \forall i\label{subeq7}\\
    &\alpha_l(R^2-\zeta_l-\norm{\mathbf{x}_l-\mathcal{O}}_2^2)=0, \forall l\label{subeq8}
\end{align}
\end{subequations}
Using Eqs. \ref{subeq3} and \ref{subeq4}, and also Eqs. \ref{subeq5} and \ref{subeq6}, one concludes that $\gamma_i=0$, $\forall i$ and $\gamma_l=0$, $\forall l$. As a result, the Lagrangian in Eq. \ref{L2} would be
\begin{align}
    \nonumber \mathcal{L}&=(c_1p)^{\frac{-1}{p-1}}(1/p-1)\norm{\boldsymbol\alpha_T}_{p/(p-1)}^{p/(p-1)}+(c_2p)^{\frac{-1}{p-1}}(1/p-1)\norm{\boldsymbol\alpha_N}_{p/(p-1)}^{p/(p-1)}\\
    \nonumber&+\sum_i\alpha_i\mathbf{x}_i^\top\mathbf{x}_i-\sum_l\alpha_l\mathbf{x}_l^\top\mathbf{x}_l-\sum_i\sum_j\alpha_i\alpha_j\mathbf{x}_i^\top\mathbf{x}_j-\sum_l\sum_m\alpha_l\alpha_m\mathbf{x}_l^\top\mathbf{x}_m\\
    &+2\sum_i\sum_l\alpha_i\alpha_l\mathbf{x}_i^\top\mathbf{x}_l
    \label{L3}
\end{align}
The dual problem then reads
\begin{align}
    \nonumber &\max_{\boldsymbol\alpha_T,\boldsymbol\alpha_N} \hspace{10pt}\mathcal{L}\\
    \nonumber\text{s.t. }& \boldsymbol\alpha_T\geq0,\boldsymbol\alpha_N\geq0,\\ &\norm{\boldsymbol\alpha_T}_1-\norm{\boldsymbol\alpha_N}_1=1
    \label{dual2}
\end{align}
Since $p\geq1$ leads to $p/(p-1)\geq1$, the terms $\norm{.}_{p/(p-1)}^{p/(p-1)}$ in the Lagrangian are convex while the remaining terms are either linear or quadratic functions and the constraint sets are affine. Subsequently, the maximisation problem in Eq. \ref{dual2} is convex.
\subsection{Joint formulation}
As discussed earlier, when only positive labelled training observations are available, in the proposed approach one solves the optimisation problem in Eq. \ref{optL1} with the Lagrangian given in Eq. \ref{L1}. When in addition to the positive training samples, labelled negative training objects are also available, the problem to be solved is expressed as the optimisation task in Eq. \ref{dual2} with the corresponding Lagrangian given in Eq. \ref{L3}. Although the optimisation tasks corresponding to the pure positive case and that of the second scenario where negative training samples are also available may appear different, nevertheless, both optimisation problems can be expressed compactly using a joint formulation as follows. Let us assume that vector $\mathbf{y}$ corresponds to the labels of training samples where for positive objects the label is $+1$ while for any possible non-target training samples the corresponding label is $-1$. Furthermore, suppose the Lagrange multipliers associated with the negative and positive samples are all collected into a single vector $\boldsymbol\alpha$. In order to reduce the clutter in the formulation, let us further assume $q=p/(p-1)$, $\bar{c_1}=\frac{1}{2}(c_1p)^{\frac{-1}{p-1}}(1-1/p)$ and $\bar{c_2}=\frac{1}{2}(c_2p)^{\frac{-1}{p-1}}(1-1/p)$. With these definitions, the Lagrangian in Eq. \ref{L3} may be expressed as
\begin{align}
    \mathcal{L}=-\bar{c_1}\norm{\boldsymbol\alpha\odot(1+\mathbf{y})}_q^q-\bar{c_2}\norm{\boldsymbol\alpha\odot(1-\mathbf{y})}_q^q+\sum_i\alpha_iy_i\mathbf{x}_i^\top\mathbf{x}_i-\sum_{i,j}\alpha_iy_i\alpha_jy_j\mathbf{x}_i^\top\mathbf{x}_j
    \label{JL}
\end{align}
\noindent where $\odot$ denotes hadamard/elementwise product. It may be easily verified that when only positive training samples are available, the Lagrangian above correctly recovers that of Eq. \ref{L1} while in the existence of labelled negative training objects, it matches that of Eq. \ref{L3}. As a result, in the proposed approach, the generic optimisation problem to solve can be expressed as
 \begin{align}
    \nonumber &\max_{\boldsymbol\alpha} \hspace{10pt}\mathcal{L}\\
    \text{s.t. }& \boldsymbol\alpha\geq0,\mathbf{y}^\top\boldsymbol\alpha=1
\end{align}
\noindent where $\mathbf{y}$ is the vectors of labels and the Lagrangian $\mathcal{L}$ is given as Eq. \ref{JL}.
\subsection{Kernel space representation}
In may practical applications, instead of a rigid boundary, a more elastic description is favoured. In such cases, a reproducing kernel Hilbert space representation may be adopted. Inspecting the Lagrangian in Eq. \ref{JL}, it can be observed that the training samples only appear in terms of inner products which facilitates deriving a kernel-space representation for the proposed approach. Since in the kernel space it holds that $\phi(\mathbf{x}_i)^\top\phi(\mathbf{x}_j)=\kappa(\mathbf{x}_i,\mathbf{x}_j)$ where $\kappa(.,.)$ is the kernel function, the Lagrangian in the reproducing kernel Hilbert space may be written as
\begin{align}
    \mathcal{L}=-\bar{c_1}\norm{\boldsymbol\alpha\odot(1+\mathbf{y})}_q^q-\bar{c_2}\norm{\boldsymbol\alpha\odot(1-\mathbf{y})}_q^q+\sum_i\alpha_iy_i\kappa(\mathbf{x}_i,\mathbf{x}_i)-(\boldsymbol\alpha\odot\mathbf{y})^\top\mathbf{K}(\boldsymbol\alpha\odot\mathbf{y})
\end{align}
\noindent where $\mathbf{K}$ denotes the kernel matrix. If additionally, all objects have unit length in the feature space , i.e. if $\phi(\mathbf{x}_i)^\top\phi(\mathbf{x}_i)=1$, one may further simplify the Lagrangian. For this propose, note that as for normalised feature vectors we have $\sum_i\alpha_iy_i\kappa(\mathbf{x}_i,\mathbf{x}_i)=\mathbf{y}^\top\boldsymbol\alpha$ and since due to the constraints imposed it must hold that $\mathbf{y}^\top\boldsymbol\alpha=1$, the term $\sum_i\alpha_iy_i\kappa(\mathbf{x}_i,\mathbf{x}_i)$ can be safely dropped from the objective function without affecting the result. As a result, the optimisation problem for unit-length features shall be
\begin{align}
    \nonumber \min_{\boldsymbol\alpha} \hspace{10pt}\bar{c_1}\norm{\boldsymbol\alpha\odot(1+\mathbf{y})}_q^q&+\bar{c_2}\norm{\boldsymbol\alpha\odot(1-\mathbf{y})}_q^q+(\boldsymbol\alpha\odot\mathbf{y})^\top\mathbf{K}(\boldsymbol\alpha\odot\mathbf{y})\\
    &\text{s.t. } \boldsymbol\alpha\geq0,\mathbf{y}^\top\boldsymbol\alpha=1
    \label{1L}
\end{align}
As a widely used kernel function, the Gaussian kernel by definition, yields unit-length feature vectors in the kernel space, and the formulation above is applicable.
\subsection{Decision strategy}
Similar to the conventional SVDD approach, for decision making in the proposed $\ell_p$ slack norm method, the distance of an object to the centre of the description is gauged and employed as a dissimilarity criterion. The distance of an object $\mathbf{z}$ to the centre of the hypersphere $\mathcal{O}$ in the kernel space is
\begin{align}
    f(\mathbf{z})=\norm{\phi(\mathbf{z})-\phi(\mathcal{O})}_2^2=\kappa(\mathbf{z},\mathbf{z})-2\sum_i\alpha_iy_i\kappa(\mathbf{z},\mathbf{x}_i)+(\boldsymbol\alpha\odot\mathbf{y})^\top\mathbf{K}(\boldsymbol\alpha\odot\mathbf{y})
    \label{distf}
\end{align}
In order to compute the radius of the description, note that the complementary conditions in Eqs. \ref{subeq7} and \ref{subeq8} may be compactly represented as $\alpha_j(R^2+y_j\zeta_j-\norm{\phi(\mathbf{x}_j)-\phi(\mathcal{O})}_2^2)=0$. As a result, if for an object $\mathbf{x}_j$ the corresponding Lagrange multiplier $\alpha_j$ is non-zero, it must hold that $R^2+y_j\zeta_j-\norm{\phi(\mathbf{x}_j)-\phi(\mathcal{O})}_2^2=0$, and hence, the radius of the description may be computed as
\begin{align}
    \nonumber R^2 &= \norm{\phi(\mathbf{x}_j)-\phi(\mathcal{O})}_2^2-y_j\zeta_j\\
    &=\kappa(\mathbf{x}_j,\mathbf{x}_j)-2\sum_i\alpha_iy_i\kappa(\mathbf{x}_j,\mathbf{x}_i)+(\boldsymbol\alpha\odot\mathbf{y})^\top\mathbf{K}(\boldsymbol\alpha\odot\mathbf{y})-y_j\zeta_j
\end{align}
\noindent where $j$ indexes an object whose corresponding Lagrange multiplier $\alpha_j$ is non-zero. The objects whose distance to the centre of the hyper-sphere is larger than the radius (subject to some margin) would be classified as novel.

\section{Generalisation error bound}
\label{geb}
In this section, using the Rademacher complexities, we characterise the generalisation error bound for the proposed $\ell_p$ slack norm SVDD approach.
\begin{thm}
Let us assume $\mathcal{F}$ corresponds to a class of kernel-based linear functions:
\begin{align}
    \mathcal{F}=\{x\rightarrow \mathbf{w}^\top\phi(x), \norm{\mathbf{w}}_2\leq B\}
\end{align}
then the empirical Rademacher complexity of function class $\mathcal{F}$ over samples $(\mathbf{x}_i)_{i=1}^n$, denoted as $\hat{\mathcal{R}}_n(\mathcal{F})$, is bounded as \cite{shawe-taylor_cristianini_2004}
\begin{align}
    \hat{\mathcal{R}}_n(\mathcal{F})\leq\frac{2B}{n}\sqrt{tr(\mathbf{K})}\leq\frac{2BB_\kappa}{\sqrt{n}}
\end{align}
where $tr(.)$ denotes matrix trace and $\mathbf{K}$ stands for the kernel matrix associated with the feature mapping $\phi(.)$ and $B_\kappa^2$ is an upper bound on the kernel function $\kappa(.,.)$.
\label{KCRC}
\end{thm}
Next, we present the main theorem concerning the generalisation performance of the proposed approach.
\begin{thm}
In the proposed approach, assuming that $\upsilon$ is a margin parameter, with confidence greater than $1-\Delta$, a test point $\mathbf{x}$ is incorrectly classified with the probability bounded as
\begin{align}
    P[y\big(f(\mathbf{x})-R^2\big)>\upsilon]\leq \frac{1}{n\upsilon^p}\norm{\boldsymbol\zeta}_p^p+\frac{4pBB_\kappa}{\upsilon^p\sqrt{n}}(B^2+3B_\kappa^2+R^2)^{p-1}+3\sqrt{\frac{\ln(2/\Delta)}{2n}}
    \label{PoE}
\end{align}
\noindent where $y$ is the ground truth label for observation $\mathbf{x}$.
\label{EP}
\end{thm}
For the proof of Theorem \ref{EP}, first, we review a few relevant theories and then present the proof.
\begin{thm}
Assume $\delta\in(0,1)$ and suppose $\mathcal{G}$ is a function class from $X$ to $[0,1]$. Let $(\mathbf{x}_i)_{i=1}^n$ be independent samples that are drawn according to a probability distribution $\mathcal{D}$. Then with a probability higher than $1-\Delta$ over $(\mathbf{x}_i)_{i=1}^n$, for each $g\in\mathcal{G}$ it holds that \cite{shawe-taylor_cristianini_2004}
\begin{align}
    \mathbb{E}_\mathcal{D}[g(\mathbf{x})]\leq \hat{\mathbb{E}}[g(\mathbf{x})]+\hat{\mathcal{R}}_n(\mathcal{G})+3\sqrt{\frac{\ln(2/\Delta)}{2n}}
\end{align}
\label{generic}
\end{thm}
\noindent where $\hat{\mathbb{E}}[g(\mathbf{x})]$ is the empirical expectation of $g(\mathbf{x})$ on the random sample set $(\mathbf{x}_i)_{i=1}^n$ and $\hat{\mathcal{R}}_n(\mathcal{G})$ denotes the empirical Rademacher complexity of the function class $\mathcal{G}$.

\begin{thm}
If $\mathcal{A}:\mathbb{R}\rightarrow\mathbb{R}$ is $L$-Lipschitz and satisfies $\mathcal{A}(0)=0$, then the empirical Rademacher complexity of the composition function class $\mathcal{A}\circ\mathcal{F}$ satisfies $\hat{\mathcal{R}}_n(\mathcal{A}\circ\mathcal{F})\leq 2L\hat{\mathcal{R}}_n(\mathcal{F})$ \cite{shawe-taylor_cristianini_2004}.
\label{TrRM}
\end{thm}
Towards the proof of Theorem \ref{EP}, we present the following theorem.
\begin{thm}
Let us consider $h(\mathbf{x})$ as the hypothesis function defined as $h(\mathbf{x})=f(\mathbf{x})-R^2$ where $f(\mathbf{x})$ measures the distance of sample $\mathbf{x}$ with label $y$ to the centre of the hypersphere in the feature space (see Eq. \ref{distf}). For some fixed margin $\upsilon>0$, we define $g(.)$ as
\begin{align}
    g(\mathbf{x})=\mathcal{A}(yh(\mathbf{x}))=    \begin{cases}
      0 & \text{if $yh(\mathbf{x})\leq 0$;}\\
      \big(yh(\mathbf{x})/\upsilon\big)^p & \text{if $0\leq yh(\mathbf{x})\leq \upsilon$;}\\
      1 & \text{else.}
    \end{cases}
\end{align}
$\mathcal{A}: \mathbb{R} \rightarrow [0,1]$ is $L$-Lipschitz and satisfies $\mathcal{A}(0)=0$. Then with a probability higher than $1-\Delta$ over $(\mathbf{x}_i)_{i=1}^n$ it holds
\begin{align}
    \mathbb{E}_\mathcal{D}[g(\mathbf{x})]\leq  \frac{1}{\upsilon^p n}\norm{\boldsymbol\zeta}_p^p+\frac{4Bp}{n\upsilon^p}(B^2+3B_{\kappa}+R^2)^{p-1}\sqrt{tr(\mathbf{K})}+3\sqrt{\frac{\ln(2/\Delta)}{2n}}
\end{align}
\label{GE}
\end{thm}
\noindent \textbf{Proof}\\
We have
\begin{align}
    \hat{\mathbb{E}}[g(\mathbf{x})]=\frac{1}{n}\sum_ig(\mathbf{x}_i)\leq\frac{1}{n\upsilon^p}\sum_i\big(y_i(f(\mathbf{x}_i)-R^2)\big)^p_+=\frac{1}{n\upsilon^p}\sum_i\zeta_i^p=\frac{1}{n\upsilon^p}\norm{\boldsymbol\zeta}_p^p
    \label{TEr}
\end{align}
where $
    (z)_+=    \begin{cases}
      0 & \text{if $z<0$}\\
      z & \text{otherwise.}
    \end{cases}
$
and $\boldsymbol\zeta$ stands for a vector collection of all $\zeta_i$'s. Note that $\mathcal{A}(.)$ is Lipschitz with constant $L$. As with the zero-one loss, the margin loss above penalises any misclassified objects but also penalises $h$ when it correctly classifies an object with low confidence. In order to derive an upper bound on $L$, observe that $\frac{\partial \mathcal{A}}{\partial \big(yh(\mathbf{x})\big)}=\frac{p}{\upsilon^p}\big(y(f(\mathbf{x})-R^2)\big)^{p-1}$, and consequently, we have
\begin{align}
    \norm{\frac{\partial \mathcal{A}}{\partial \big(yh(\mathbf{x})\big)}}_2=\frac{p}{\upsilon^p}\norm{(f(\mathbf{x})-R^2)}_2^{p-1}\leq\frac{p}{\upsilon^p}(\norm{f(\mathbf{x})}_2+R^2)^{p-1}
\end{align}
Since the kernel function is bounded by $B_{\kappa}^2$, using Eq. \ref{distf}, and the fact that $\norm{\mathbf{w}}_2^2=\boldsymbol\alpha^\top\mathbf{K}\boldsymbol\alpha$ \cite{shawe-taylor_cristianini_2004}, we have
\begin{align}
    \norm{f(\mathbf{x})}_2\leq B^2+3B_{\kappa}^2
\end{align}
and hence
\begin{align}
    \norm{\frac{\partial \mathcal{A}}{\partial \big(yh(\mathbf{x})\big)}}_2\leq\frac{p}{\upsilon^p}(B^2+3B_{\kappa}^2+R^2)^{p-1}
\end{align}
As a result, $\mathcal{A(.)}$ is Lipschitz with constant $L=\frac{p}{\upsilon^p}(B^2+3B_{\kappa}^2+R^2)^{p-1}$.

Next, using Theorem \ref{TrRM} and Theorem \ref{KCRC}, we have
\begin{align}
    \hat{\mathcal{R}}_n(\mathcal{G})\leq 2L\hat{\mathcal{R}}_n(\mathcal{F})\leq\frac{4BB_\kappa L}{\sqrt{n}}\leq\frac{4pBB_\kappa}{\upsilon^p\sqrt{n}}(B^2+3B_\kappa^2+R^2)^{p-1}
    \label{ERC}
\end{align}
Using Eq. \ref{TEr} and Eq. \ref{ERC} in Theorem \ref{generic}, the proof to Theorem \ref{GE} is complete. Since we have $P[y\big(f(\mathbf{x})-R^2\big)>\upsilon]\leq \mathbb{E}_\mathcal{D}[g(\mathbf{x})]$, using Theorem \ref{GE}, the proof to Theorem \ref{EP} is completed. $\square$

As may be observed from Eq. \ref{PoE}, parameter $p$ directly affects the expected loss on the training set (the first term on the RHS of the equation) and also controls the Rademacher complexity (the second term on the RHS of Eq. \ref{PoE}) of the proposed method. As the error probability varies as a function of $p$, the utility of a free norm parameter in the proposed approach is justified. Note that depending on $\boldsymbol\zeta$ and the margin parameter $\upsilon$, setting $p=1$ may not minimise the RHS in Eq. \ref{PoE}, and hence, may lead to an increased probability of misclassification in the proposed approach. In practice, the norm parameter $p$ may be adjusted according to the characteristics of the data using cross validation to optimise the performance or to control the false acceptance/rejection rate. Note also, since parameter $p$ appears in the dual problem as $\norm{.}^{p/(p-1)}_{p/(p-1)}$ terms (see Eq. \ref{1L}), it also affects the sparsity of $\boldsymbol\alpha$.

\section{Experiments}
\label{exps}
In this section, an experimental evaluation of the proposed approach is conducted where we provide a comparison to some other variants of the SVDD approach as well as to baseline approaches on multiple datasets. The rest of this section is organised as detailed next.
\begin{itemize}
\item In Section \ref{bound}, we visualise the decision boundaries inferred by the proposed approach for different $p$'s for synthetic data.
\item In Section \ref{dets}, the implementation details, the experimental set-up, and the standard datasets used in the experiments are discussed.
\item In Section \ref{puree}, the results of an experimental evaluation of the proposed approach in a pure one-class setting (labelled negative objects unavailable) are presented and compared with other methods on multiple datasets.
\item Section \ref{negg} provides the results of an experimental evaluation of the proposed approach in the presence of negative training samples along with a comparison against other methods on multiple datasets.
\end{itemize}

\subsection{Decision boundaries}
\label{bound}
\begin{figure}[t]
\centering
\includegraphics[scale=.5]{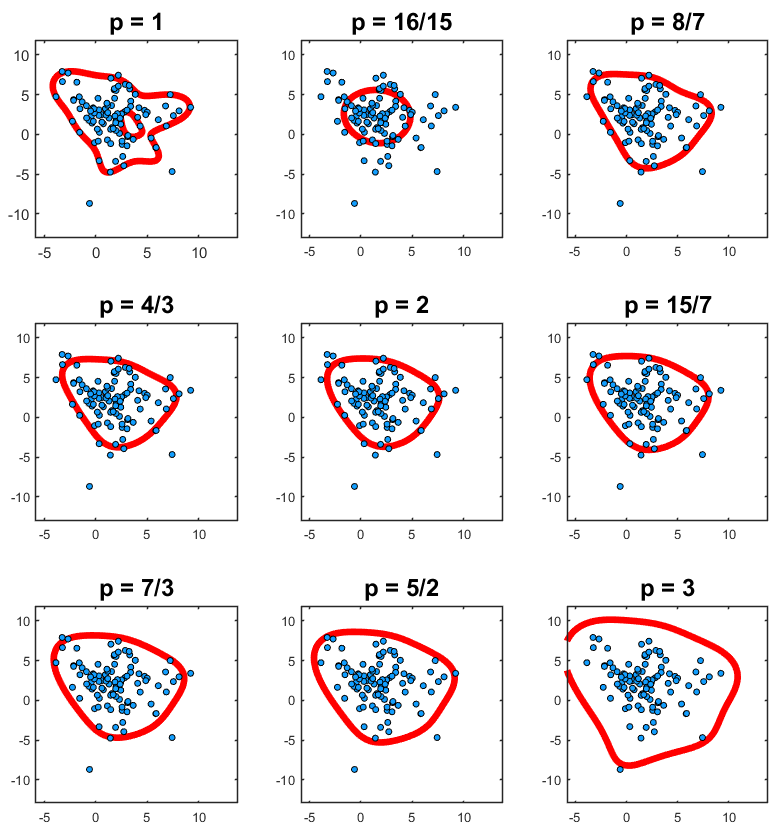}
\caption{Decision boundaries for the proposed $\ell_p$-SVDD approach with a Gaussian kernel function for different values of $p$ for $100$ normally distributed random samples with mean of $2$ and standard deviation of $3$ in each dimension. ($p = 1$ corresponds to the original SVDD method).}
\label{boundaries}
\end{figure}

In order to visualise the effect of norm parameter $p$ on the inferred decision boundaries, we randomly generate 100 normally distributed 2D samples with a mean of 2 and standard deviation of 3 in each direction. Using a Gaussian kernel function, the proposed approach is then run to derive a description of the data. The experiment is repeated for different values of $p\in\{1,16/15,8/7,4/3,2,15/7,7/3,5/2,3 \}$ where $p=1$ corresponds to the original SVDD method in \cite{Tax2004}. The decision boundaries superimposed on the data are visualised in Fig. \ref{boundaries}. From the figure, it may be observed that for the case of $p=1$ the method has inferred a boundary which separates a region of relatively low density in the middle of the distribution from the rest of the 2D space. For the random data samples generated in this experiment with a mean of $(2,2)$ this clearly indicates a case of over-fitting. By increasing the norm parameter above 1, the decision boundary better covers the mean of the distribution. More specifically, while for smaller values of $p$ the boundary is tighter, for larger values the description tends to encapsulate a higher percentage of the normal samples. As will be discussed in the following sections, in the proposed method, we tune parameter $p$ using the validation set corresponding to each dataset.

\subsection{Implementation details}
\label{dets}
In the experiments that follow, the features are first standardised by subtracting the mean computed over all positive training samples and then dividing by the standard deviation followed by normalising each feature vector to have a unit $\ell_2$-norm. The positive samples are divided randomly into three non-overlapping subsets to form the training, validation, and the test sets. Similarly, the negative samples are divided randomly into three disjoint subsets for training, validation and testing purposes. In order to minimise possible effects of random data partitioning on the performance, we repeat the procedure above 10 times, and record the mean along with the standard deviation of the performance over these 10 trials. We set the parameters of all methods on the corresponding validation subset of each dataset. In particular, for the proposed approach $p\in\{32/31, 16/15, 8/7, 6/5, 4/3, 3/2, 2, 5/2, 5, 20\}$ and $c_1,c_2\in \{10^{-3},10^{-2},10^{-1},1\}$. In all experiments, we use a Gaussian kernel the width of which is set to half of the average pairwise Euclidean distance among all training objects. As the dual problem in Eq. \ref{1L} is convex, one may use different algorithms for optimisation. In this work, we use CVX \cite{cvx}, a package for solving convex programmes.
\begin{table}
\renewcommand{\arraystretch}{1.2}
\caption{Statistics of the datasets used.}
\label{mat_inv}
\centering
\footnotesize{
\begin{tabular}{llcccc}
\hline
Abbrev.&Dataset & \#total & \#positive & dim. & Source\\
\hline
D1&Iris (virginica) &150&50&4&UCI\\
D2&Hepatitis (normal) &155&123&19&UCI\\
D3&Ecoli (periplasm) &336&52&7&UCI\\
D4&Concordia16 (digit 1)&4000&400&256&CENPARMI\\
D5&Delft pump (5x1)&336&133&64&Delft\\
D6&Balance-scale (middle)&625&49&4&UCI\\
D7&Wine (2)&178&71&13&UCI\\
D8&Waveform (0)&900&300&21&UCI\\
D9&Survival ($<$5yr)&306&81&3&UCI\\
D10&Housing (MEDV$>$35)&506&48&13&Statlib\\
D11&glass6&214&185&9&KEEL\\
D12&haberman&306&225&3&KEEL\\
D13&led7digit&443&406&7&KEEL\\
D14&pima&768&500&8&KEEL\\
D15&wisconsin&683&444&9&KEEL\\
D16&yeast(0-5-6-7-9-vs-4)&528&477&8&KEEL\\
D17&cleveland(0-vs-4)&177&164&13&KEEL\\
D18&Breast(benign)&699&458&9&UCI\\
D19&Survival ($>$5yr)&306&225&3&UCI\\
D20&FMNIST (Class "1")&1926&1027&784&Zalando\\
\hline
\label{dstats}
\end{tabular}}
\end{table}

In order to evaluate the proposed approach, 20 benchmark databases from the UCI repository \cite{ucirep}, TUDelf University \cite{delftrep}, KEEL repository \cite{keelrep}, CENPARMI \cite{concordiarep}, Statlib \cite{HARRISON197881}, and Zalando \cite{xiao2017/online} are used. The datasets used in the experiments correspond to different application domains from varied sources. The statistics of the datasets used in the experiments are reported in Table \ref{dstats}. For the evaluation of the proposed approach, we conduct two sets of experiments. The first set follows a pure one-class classification paradigm, i.e. only positive samples are used to train the models. In the second set of experiments, negative objects are also deployed for model training. For comparison, we report the performance of the original SVDD approach of \cite{Tax2004} denoted as "$\ell_1$-SVDD" and also its alternative variant which considers squared errors in the objective function, denoted as "$\ell_2$-SVDD" \cite{Chang_arevisit}. The proposed approach is denoted as "$\ell_p$-SVDD" in the corresponding tables. We also provide a comparison of the proposed $\ell_p$-SVDD method to some linear re-weighting variants of the SVDD approach including P-SVDD \cite{WANG2013875}, DW-SVDD \cite{CHA20143343}, and GL-SVDD \cite{9632278} as well as state-of-the-art OCC techniques. In particular, we have included kernel-based one-class classifiers which are applicable to moderately-sized datasets. These are the kernel Gaussian Process method (GP) \cite{KEMMLER20133507}, the Kernel Null Foley-Sammon Transform (KNFST) \cite{4721915,9055448}, and the Kernel Principal Component Analysis (KPCA) \cite{HOFFMANN2007863}.

Following the common approach in the literature and in order to facilitate the comparison of the performances of different methods independent from a specific operating threshold, we report the performances in terms of the AUC measure which is the area under the Receiver Operating Characteristic curves (ROC). The ROC curve characterises the true positive rate against the false positive rate at various operating thresholds. A higher AUC indicates a better performance for the system.
\begin{sidewaystable}
\renewcommand{\arraystretch}{1.2}
\caption{Comparison of the performance of OCC approaches on different datasets in a pure one-class scenario in terms of $\%$AUC.}
\label{comps1}
\centering
\footnotesize{
\begin{tabular}{lccccccccc}
\hline
 &GP&KPCA&KNFST&P-SVDD&DW-SVDD&GL-SVDD& $\ell_1$-SVDD & $\ell_2$-SVDD & $\ell_p$-SVDD\\
\hline
D1 &68.03$\pm$16.51 &78.17$\pm$22.69 &68.51$\pm$16.46 &78.98$\pm$15.58 &67.42$\pm$16.75 &67.70$\pm$16.39 &67.42$\pm$16.75 &67.37$\pm$16.52 &$\mathbf{81.23\pm11.26}$\\
D2 &58.78$\pm$4.37 &66.46$\pm$6.75 &58.65$\pm$4.39 &62.51$\pm$5.98 &59.36$\pm$4.37 &61.20$\pm$6.06 &59.98$\pm$4.31 &59.48$\pm$4.26 &$\mathbf{66.60\pm6.80}$\\
D3 &61.30$\pm$11.33 &53.06$\pm$13.66 &61.32$\pm$11.15 &53.98$\pm$12.31 &61.67$\pm$11.66 &61.08$\pm$10.49 &61.08$\pm$11.36 &61.20$\pm$11.46 &$\mathbf{62.19\pm10.53}$\\
D4 &92.96$\pm$2.60 &94.10$\pm$1.73 &92.94$\pm$2.62 &93.24$\pm$2.50 &93.31$\pm$2.54 &93.21$\pm$2.53 &94.36$\pm$1.84 &93.58$\pm$2.42 &$\mathbf{94.43\pm1.58}$\\
D5 &87.87$\pm$3.43 &86.07$\pm$3.29 &87.85$\pm$3.45 &84.78$\pm$3.92 &87.85$\pm$3.45 &88.02$\pm$3.33 &87.85$\pm$3.45 &87.90$\pm$3.40 &$\mathbf{88.41\pm2.77}$\\
D6 &94.52$\pm$3.57 &90.77$\pm$2.46 &94.66$\pm$3.52 &92.94$\pm$3.16 &94.68$\pm$3.54 &94.64$\pm$3.54 &94.70$\pm$3.55 &94.60$\pm$3.56 &$\mathbf{94.79\pm3.59}$ \\
D7 &60.82$\pm$11.03 &72.30$\pm$11.10 &60.71$\pm$10.99 &65.64$\pm$9.04 &61.12$\pm$11.57 &64.12$\pm$10.85 &60.79$\pm$11.04 &60.90$\pm$11.19 &$\mathbf{72.53\pm11.36}$\\
D8 &64.15$\pm$4.66 &61.47$\pm$1.90 &64.10$\pm$4.64 &64.09$\pm$5.91 &65.37$\pm$5.09 &65.39$\pm$5.41 &64.91$\pm$5.53 &65.10$\pm$5.48 &$\mathbf{65.86\pm4.23}$\\
D9 &53.41$\pm$18.99 &47.37$\pm$14.10 &59.66$\pm$16.34 &52.29$\pm$19.26 &58.28$\pm$21.15 &66.40$\pm$12.11 &64.43$\pm$14.87 &60.72$\pm$15.26 &$\mathbf{67.40\pm10.21}$\\
D10 &87.73$\pm$6.00 &78.70$\pm$8.49 &87.76$\pm$6.02 &81.28$\pm$11.89 &87.77$\pm$6.11 &87.82$\pm$6.03 &87.77$\pm$6.11 &87.78$\pm$6.08 &$\mathbf{87.91\pm5.92}$\\
D11 &87.81$\pm$10.11 &95.44$\pm$2.59 &87.55$\pm$10.65 &95.48$\pm$3.53 &86.90$\pm$10.06 &96.58$\pm$2.18 &94.24$\pm$2.53 &90.68$\pm$5.67 &$\mathbf{96.73\pm1.38}$\\
D12 &59.70$\pm$5.90 &70.66$\pm$5.79 &55.54$\pm$5.51 &59.75$\pm$7.79 &55.84$\pm$7.12 &$\mathbf{70.77\pm5.96}$&58.34$\pm$7.41 &66.02$\pm$4.94 &$\mathbf{70.77\pm5.88}$ \\
D13 &62.63$\pm$18.55 &67.49$\pm$7.79 &62.34$\pm$18.01 &66.60$\pm$15.73 &41.54$\pm$6.43 &66.77$\pm$14.25 &62.84$\pm$13.62 &69.04$\pm$10.99 &$\mathbf{69.41\pm9.62}$\\
D14 &51.66$\pm$5.36 &$\mathbf{71.43\pm3.41}$&51.46$\pm$5.32 &53.47$\pm$5.42 &54.93$\pm$4.78 &56.04$\pm$4.46 &61.24$\pm$4.95 &59.83$\pm$3.75 &$\mathbf{71.43\pm3.57}$\\
D15 &52.70$\pm$12.37 &95.88$\pm$1.05 &53.51$\pm$12.56 &65.21$\pm$12.28 &47.10$\pm$14.97 &70.68$\pm$10.68 &93.39$\pm$1.73 &68.75$\pm$10.83 &$\mathbf{95.91\pm1.03}$\\
D16 &57.74$\pm$9.93 &63.71$\pm$7.39 &58.56$\pm$8.80 &62.94$\pm$8.48 &61.02$\pm$10.33 &63.26$\pm$9.98 &60.93$\pm$9.65 &61.52$\pm$10.68 &$\mathbf{64.23\pm8.29}$\\
D17 &51.49$\pm$13.99 &79.13$\pm$9.71 &51.78$\pm$14.04 &57.93$\pm$13.99 &51.09$\pm$13.90 &51.85$\pm$14.47 &51.09$\pm$13.90 &51.35$\pm$14.06 &$\mathbf{79.38\pm9.27}$\\
D18 &$\mathbf{45.02\pm3.25}$&38.11$\pm$2.53 &45.74$\pm$2.93 &41.15$\pm$5.13 &42.64$\pm$5.04 &42.07$\pm$5.14 &41.22$\pm$5.62 &41.99$\pm$5.16 &41.76$\pm$5.67 \\
D19 &55.16$\pm$13.36&37.01$\pm$8.20&44.70$\pm$11.81&51.49$\pm$13.02&61.38$\pm$14.80&61.01$\pm$9.08&60.28$\pm$9.47&58.64$\pm$11.56&$\mathbf{62.60\pm9.83}$\\
D20&94.82$\pm$1.47&94.86$\pm$10.9&94.78$\pm$1.49&94.96$\pm$1.23&95.37$\pm$0.87&$\mathbf{96.92\pm0.73}$&96.37$\pm$0.79&96.48$\pm$1.14&$\mathbf{96.92\pm0.63}$\\
\hline
\end{tabular}}
\end{sidewaystable}
\subsection{Pure one-class setting}
\label{puree}
In this setting, only positive objects are used for training. Table \ref{comps1} reports the performances of different methods in this setting where we set parameter $p$ on the validation subset of each dataset to maximise the performance. A number of observations from the table are in order. First, on all datasets the proposed $\ell_p$-SVDD approach yields a superior performance compared to its $\ell_1$-SVDD and $\ell_2$-SVDD variants. In particular, on some datasets such as D1 and D14, the improvement in the performance offered by the proposed approach is substantial while on some other datasets such as D17 the improvement is huge and reaches $28\%$. It should be noted that the performance improvements offered by the proposed approach are obtained despite the fact that the validation sets of some datasets may not be very large, and hence, may not serve as a very good representative of the entire the distribution of samples for tuning parameter $p$. It is expected that a more representative validation set would lead to even further improvements in the performance. 
A statistical ranking of different methods in the pure one-class setting using the Friedman’s test is provided in Table \ref{rank1}. From the table, it can be observed that while the proposed $\ell_p$-SVDD approach ranks the best among other approaches while the standard $\ell_1$-SVDD approach ranks much worst which underlines the significance of the proposed $\ell_p$ slack norm approach. Furthermore, although the $\ell_2$-SVDD method provides some improvement with respect to the original $\ell_1$-SVDD approach, its performance is still inferior compared to the proposed method. The second best performing method (in terms of average ranking) corresponds to a sample re-weighting SVDD approach presented in \cite{9632278} which uses global and local statistics to linearly weight slacks in the objective function.

\begin{table}
\footnotesize
\renewcommand{\arraystretch}{1.2}
\caption{Average ranking of different OCC methods in a pure one-class setting using the Friedman's test. (p-value=$5.65e-10$)}
\label{rank1}
\centering
\begin{tabular}{lc}
\hline
 \textbf{Algorithm}&\textbf{Rank}\\
 \hline
GP&6.50\\
KPCA&5.30\\
KNFST&6.70\\
P-SVDD&5.85\\
DW-SVDD&5.72\\
GL-SVDD&3.65\\
$\ell_1$-SVDD&5.20\\
$\ell_2$-SVDD& 4.80\\
$\ell_p$-SVDD (this work)&$\mathbf{1.27}$\\
\hline
\end{tabular}
\end{table}

\subsection{Training in the presence of negative data}
\label{negg}
In this second evaluation setting, in addition to positive objects, labelled negative samples are also used for training. Table \ref{comps2} reports the performances of different methods in this setting. Note that as in the case of pure one-class learning, the optimal $p$ value for the proposed approach is determined on the validation set. From among the GP, KPCA and KNFST approaches, only the KNFST approach is able to directly deploy negative samples for training. In order to emphasise that a method uses negative objects for training, a negative exponent ("$^-$") is used in the table. We also include the P-SVDD, DW-SVDD, and the GL-SVDD approaches trained using both negative and positive samples and denote them as P-SVDD$^-$, DW-SVDD$^-$, and GL-SVDD$^-$. From Table \ref{comps2}, it can be observed that on all datasets the proposed $\ell_p$-SVDD approach obtains a better performance as compared with its $\ell_1$-SVDD and $\ell_2$-SVDD variants. In particular, while on some datasets the $\ell_1$-SVDD and $\ell_2$-SVDD approaches are unable to effectively utilise negative training samples, the proposed $\ell_p$-SVDD method can better benefit from such samples to refine the description for improved performance. When compared with linear sample re-weighting methods of P-SVDD$^-$, DW-SVDD$^-$, and GL-SVDD$^-$, the proposed approach also performs better. An average ranking of different methods in this evaluation setting is provided in Table \ref{rank2}. From Table \ref{rank2} it may be seen that the proposed $\ell_p$-SVDD$^-$ approach utilising negative objects for training ranks the best among other competitors. Furthermore, neither the $\ell_1$-SVDD$^-$ nor the $\ell_2$-SVDD$^-$ methods which use negative training samples do not rank the second. The second best performing method in this setting corresponds to the KNFST method \cite{4721915,9055448}.

\begin{table}
\renewcommand{\arraystretch}{1.2}
\caption{Comparison of the performance of OCC approaches on different datasets in the presence of negative training objects in terms of $\%$AUC.}
\label{comps2}
\centering
\tiny{
\begin{tabular}{lccccccc}
\hline
\textbf{Dataset} & KNFST${^-}$&P-SVDD$^-$&DW-SVDD$^-$&GL-SVDD$^-$&$\ell_1$-SVDD${^-}$ & $\ell_2$-SVDD${^-}$ & $\ell_p$-SVDD${^-}$\\
\hline
D1 &95.90$\pm$5.48 &39.45$\pm$21.77 &74.38$\pm$24.80 &64.62$\pm$19.09 &58.53$\pm$21.63 &59.79$\pm$22.50 &$\mathbf{100.00\pm0.00}$\\
D2 &69.25$\pm$11.91 &55.88$\pm$4.63 &58.98$\pm$5.02 &66.41$\pm$6.91 &59.38$\pm$4.99 &59.60$\pm$5.12 &$\mathbf{71.57\pm7.92}$\\
D3 &70.80$\pm$4.94 &52.53$\pm$8.94 &60.70$\pm$7.55 &60.71$\pm$8.34 &59.19$\pm$8.86 &59.95$\pm$8.56 &$\mathbf{75.82\pm5.37}$\\
D4 &$\mathbf{96.56\pm1.50}$ &93.27$\pm$1.71 &94.92$\pm$1.45 &93.33$\pm$1.79 &94.46$\pm$1.23 &93.88$\pm$1.23 &$\mathbf{96.56\pm1.43}$\\
D4 &93.02$\pm$2.11 &88.63$\pm$5.07 &91.17$\pm$3.49 &91.17$\pm$3.49 &91.17$\pm$3.49 &91.19$\pm$3.45 &$\mathbf{93.32\pm1.94}$\\
D6 &90.29$\pm$12.84 &88.73$\pm$3.46 &92.61$\pm$4.15 &92.56$\pm$4.14 &92.62$\pm$4.14 &92.41$\pm$4.23 &$\mathbf{96.11\pm4.79}$\\
D7 &93.70$\pm$3.77 &50.84$\pm$8.34 &73.84$\pm$6.50 &58.59$\pm$8.39 &58.13$\pm$9.14 &58.66$\pm$8.90 &$\mathbf{94.83\pm3.00}$\\
D8 &90.01$\pm$1.31 &64.97$\pm$3.74 &66.93$\pm$3.24 &67.20$\pm$3.15 &66.95$\pm$3.30 &65.95$\pm$3.35 &$\mathbf{91.51\pm1.40}$\\
D9 &64.14$\pm$9.19 &48.56$\pm$10.65 &83.08$\pm$10.33 &63.30$\pm$11.58 &59.92$\pm$9.96 &60.72$\pm$15.15 &$\mathbf{96.44\pm2.25}$\\
D10 &89.37$\pm$7.15 &82.92$\pm$7.86 &86.66$\pm$8.53 &85.88$\pm$8.47 &86.66$\pm$8.53 &86.58$\pm$8.58 &$\mathbf{89.81\pm5.00}$\\
D11 &96.52$\pm$3.62 &84.95$\pm$11.15 &94.90$\pm$1.78 &96.50$\pm$1.86 &94.02$\pm$3.96 &89.88$\pm$5.76 &$\mathbf{97.12\pm1.72}$\\
D12 &52.52$\pm$7.84 &58.84$\pm$6.19 &63.06$\pm$5.22 &69.44$\pm$7.54 &62.10$\pm$9.37 &65.95$\pm$6.16 &$\mathbf{71.23\pm7.42}$\\
D13 &91.11$\pm$7.62 &68.97$\pm$12.57 &47.65$\pm$13.31 &69.24$\pm$13.25 &69.54$\pm$10.59 &66.53$\pm$8.41 &$\mathbf{95.20\pm2.93}$\\
D14 &63.94$\pm$3.85 &55.37$\pm$4.12 &64.88$\pm$4.01 &61.15$\pm$4.01 &59.14$\pm$4.89 &64.90$\pm$3.59 &$\mathbf{79.75\pm0.99}$\\
D15 &94.20$\pm$2.85 &68.65$\pm$4.92 &65.80$\pm$8.32 &91.91$\pm$2.71 &93.32$\pm$2.35 &88.39$\pm$3.73 &$\mathbf{98.69\pm0.54}$\\
D16 &61.74$\pm$12.47 &64.12$\pm$5.56 &63.19$\pm$6.27 &64.67$\pm$5.55 &64.33$\pm$5.08 &62.68$\pm$7.29 &$\mathbf{84.70\pm2.69}$\\
D17 &88.11$\pm$6.10 &51.27$\pm$14.69 &50.22$\pm$13.93 &51.75$\pm$14.62 &50.22$\pm$13.93 &50.36$\pm$13.67 &$\mathbf{92.80\pm3.73}$\\
D18 &56.83$\pm$3.88&49.04$\pm$1.92&51.52$\pm$2.54&51.74$\pm$3.55&52.02$\pm$2.40&47.06$\pm$4.54&$\mathbf{66.10\pm9.22}$\\
D19 &78.46$\pm$4.54&62.56$\pm$9.53&0.8586$\pm$10.37&65.30$\pm$9.56&66.69$\pm$9.38&50.44$\pm$12.80&$\mathbf{92.65\pm2.13}$\\
D20 &$\mathbf{99.25\pm0.40}$&95.83$\pm$1.40&97.72$\pm$0.99&97.55$\pm$0.79&97.03$\pm$1.15&97.24$\pm$0.96&98.07$\pm$0.86\\
\hline
\end{tabular}}
\end{table}

\begin{table}
\footnotesize
\renewcommand{\arraystretch}{1.2}
\caption{Average ranking of different OCC methods in the presence of negative training samples using the Friedman’s test). (p-value=$8.97e-14$)}
\label{rank2}
\centering
\begin{tabular}{lc}
\hline
 \textbf{Algorithm}&\textbf{Rank}\\
 \hline
KNFST$^-$&2.80\\
P-SVDD$^-$&6.350\\
DW-SVDD$^-$&4.45\\
GL-SVDD$^-$&3.90\\
$\ell_1$-SVDD$^-$&4.55\\
$\ell_2$-SVDD$^-$&4.90\\
$\ell_p$-SVDD$^-$ (this work)&$\mathbf{1.05}$\\
\hline
\end{tabular}
\end{table}

\subsection{Discussion}
\label{diss}
From a pattern classification perspective, our observations that an $\ell_1$-norm error function may not be the best choice for one-class classification are consistent with the findings in two/multi-class classification setting \cite{VAPNIK2021108018} where it is verified that error encoding mechanisms other than the widely used $\ell_1$-norm may yield better generalisation capabilities for the standard two-class SVM. Note that, while in \cite{VAPNIK2021108018} only the specific cases of $p=2$ and $p=\infty$ are analysed for the two/multi-class scenario, in this work, we formulated a generic $\ell_p$-norm error function for the one-class setting, and as demonstrated in the experimental evaluation section, the proposed approach yields better classification performance compared to the conventional $\ell_1$, $\ell_2$ and also sample re-weighting approaches for SVDD. 

It should be noted that although the proposed approach has been targeted towards learning data descriptions on medium-sized datasets, yet, it may be extended to work in conjunction with deep learning methods on large-scale datasets to benefit from the availability of such data. In this context, one possible direction for the extension of the proposed approach is to couple the training stage of the proposed $\ell_p$-SVDD approach with that of a deep network, akin to the DeepSVDD technique \cite{pmlr-v80-ruff18a}, for a joint learning of data representation and one-class description learning for enhanced performance.

While in the proposed approach all slacks are similarly penalised via a common $p$-norm cost, another possibility is to consider sample-specific non-linear penalty functions, for instance by further weighting each object according to some criteria, similar in spirit to the work in \cite{WANG2013875,CHA20143343,LEE20051768,CHEN2015129}. Finally, there exists multi-class classification problems where a test observation may potentially belong to none of the predefined classes, giving rise to an open-set classification problem \cite{6365193}. In this context, one may consider an extension of the proposed approach where multiple hyper-spheres, each corresponding to a single class are inferred.
\section{Conclusion}
\label{conc}
We generalised the SVDD approach from an $\ell_1$-norm risk to an $\ell_p$-norm ($p\geq1$) cost w.r.t. slacks, enabling formulating non-linear cost functions over errors. An $\ell_p$-norm ($p\geq1$) penalty function is a direct generalisation of the $\ell_1$ penalty term, and thus, the proposed approach presented a natural and intuitive generalisation of the SVDD method. By virtue of a dual representation of the problem, it was shown that the proposed approach corresponds to an optimisation task where sparsity-inducing $\ell_q$ norms ($q=p/(p-1)$) over the dual variable (i.e. $\boldsymbol\alpha$) (absent in the original SVDD formulation) are introduced into the objective function, allowing the algorithm to tune into the inherent sparsity of the data. A theoretical analysis of the proposed approach based on Rademacher complexities yielded an upper bound on the classification error, emphasising the dependence of the performance on the norm parameter $p$. The experimental evaluation on several standard OCC datasets, on the other hand, revealed that the proposed approach leads to improvements upon the existing $\ell_1$ and $\ell_2$ penalty functions in addition to some other linear sample re-weighting SVDD variants and also performs very favourably compared with other existing OCC methods both in a pure one-class setting and also in the case of training in the presence of labelled negative objects. Finally, we discussed several future research avenues regarding possible extensions of the proposed approach.
\section*{Acknowledgements}
This research is supported by The Scientific and Technological Research Council of Turkey (TÜBİTAK) under the grant no 121E465.
\bibliography{ref}

\end{document}